\documentclass[11pt,onecolumn, fleqn]{article}
\usepackage[T1]{fontenc}
\usepackage[utf8]{inputenc}
\usepackage{authblk}
\usepackage{lipsum}
\usepackage[margin=1.5in]{geometry}
\usepackage{amsmath}
\usepackage{epsfig}
\usepackage{subfigure}
\usepackage{float}
\usepackage{placeins}
\usepackage{graphicx}
\usepackage[nottoc]{tocbibind}
\usepackage{authblk}
\usepackage{enumitem}
\usepackage{todonotes}
\usepackage{booktabs}
\usepackage{makecell}

\usepackage[hidelinks]{hyperref}

\usepackage[backend=biber,style=nature,sorting=none]{biblatex}

\title{Bibliography management: BibTeX}

\author{Ehud Karpas,  Omri Abend, Yonatan Belinkov, Barak Lenz, Opher Lieber, Nir Ratner, Yoav Shoham, Hofit Bata, Yoav Levine, Kevin Leyton-Brown, Dor Muhlgay, Noam Rozen, Erez Schwartz, Gal Shachaf,  Shai Shalev-Shwartz, Amnon Shashua, Moshe Tenenholtz} 
\affil{AI21 Labs}

\title{MRKL Systems \\
\large A modular, neuro-symbolic architecture that combines large language models, external knowledge sources and discrete reasoning}

\begin{document}

\maketitle
\noindent
\textbf{Abstract}
\newline
Huge language models (LMs) have ushered in a new era for AI, serving as a gateway to natural-language-based knowledge tasks. Although an essential element of modern AI, LMs are also inherently limited in a number of ways. We discuss these limitations and how they can be avoided by adopting a systems approach. Conceptualizing the challenge as one that involves knowledge and reasoning in addition to linguistic processing, we define a flexible architecture with multiple neural models, complemented by discrete knowledge and reasoning modules. We describe this neuro-symbolic architecture, dubbed the Modular Reasoning, Knowledge and Language (MRKL, pronounced “miracle”) system, some of the technical challenges in implementing it, and Jurassic-X, AI21 Labs’ MRKL system implementation.

\section{Introduction}

Huge language models (LMs) such as BERT \cite{BERT}, GPT-3 \cite{GPT3}, Jurassic-1 \cite{J1}, PaLM \cite{PaLM}, and others \cite{T5, T0, EXT5, roberta, Turing-NLG}, have taken AI by storm, with the promise of serving as versatile, general-purpose foundations for many applications. Indeed, partly for this reason, they have been rebranded by some as “foundation models” \cite{foundation_models}. The term is meant broadly, covering language models as well as models that were trained on more than just text, and although such multimodal models are not the focus of this paper, there’s another reason to take the term ``language model'' with a grain of salt. While LMs indeed model syntax, and other linguistic elements, their most striking feature is that they model the world, as described by the data on which they were trained. And so really LMs serve as a textual gateway to the universe of knowledge \cite{lama, lpaqa}, and perhaps should instead be called “language and knowledge” models.

When viewed this way, it becomes clear that, despite their value, current LMs have inherent limitations. While  versatile and impressive, the output of even huge LMs is in many cases wrong, and often ridiculously so \cite{limitations}. Here is a sample output of GPT-3 on some simple queries. (To be clear, this is not a critique of GPT-3 specifically, and other LMs --- including our own Jurassic-1 --- exhibit similar silliness.) 
\newline
\begin{center}
    \includegraphics[width=12cm]{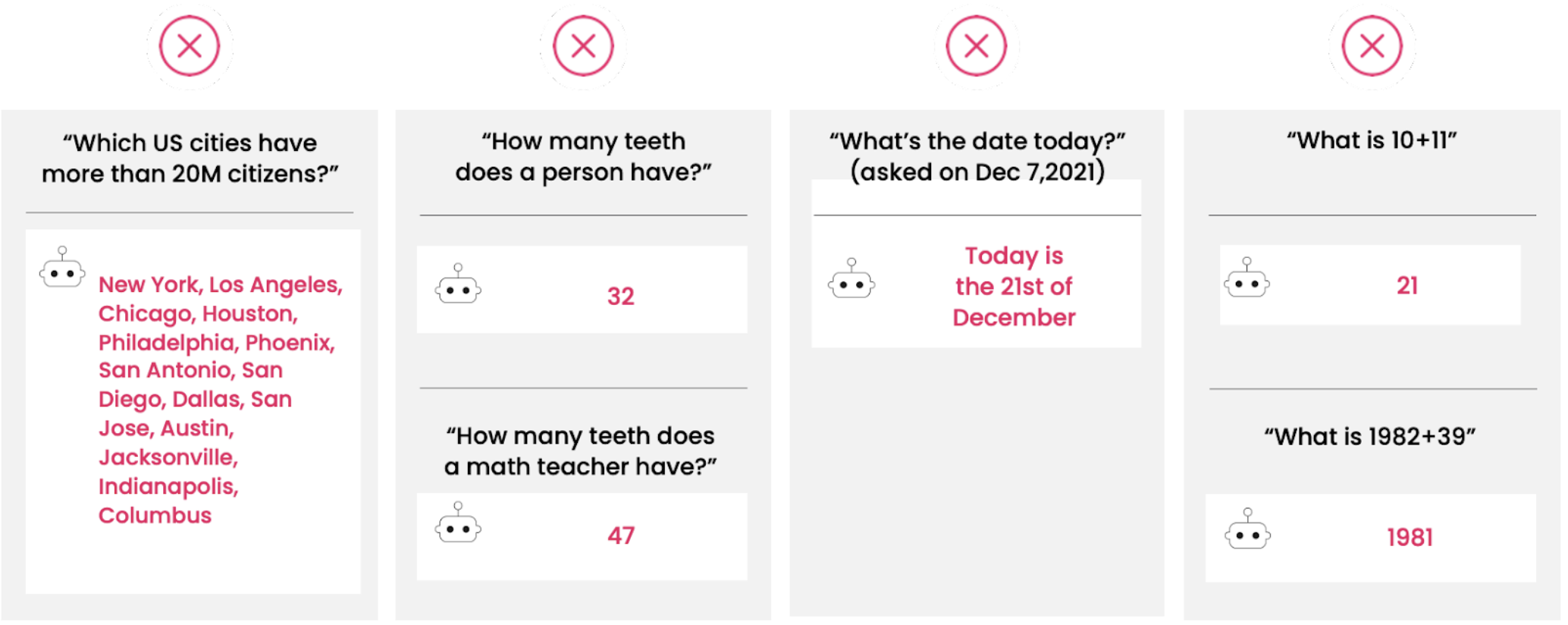}
\end{center}

For example, LMs can struggle to understand that there are no US cities with more than 20m citizens, that a math teacher is a person, don’t know what today’s date is, nor can they engage in even simple (e.g., mathematical) reasoning.

When you look for the root cause, you realize the core limitations of LMs: They don’t have access to all relevant knowledge, and neural models are ill-suited for certain types of calculation. More specifically:

\begin{enumerate}

    \item \underline{Lack of access to current information}. Certain data constantly change -- the exchange rate between the dollar and the Moroccan Dirham, current COVID numbers, the stock price of AAPL, the weather in Vancouver (OK, not so much), or even the current date. It’s impossible, by their design, for pretrained language models to keep up with this dynamic information \cite{timeLM}.

    \item \underline{Lack of access to proprietary information sources}. As an important special case of 1, the models don’t have access to proprietary  information, such as the client roster in a company’s database or the state of an online game.

    \item \underline{Lack of reasoning}. Certain reasoning is beyond the reach of the neural approach, and requires a dedicated reasoning process. We saw above the classic example of arithmetic reasoning. GPT-3 and Jurassic-1 perform well on 2-digit addition, which is impressive, but confidently spit out nonsensical answers on 4-digit additions. With increased training time, better data, and larger models, the performance of LMs will improve, but will not reach the robustness of an HP calculator from the 1970s. And mathematical reasoning is just the tip of an iceberg.

\end{enumerate}

In addition to these shortcomings, there is another inherent problem with the traditional approach to deploying LMs:

\begin{enumerate}

  \setcounter{enumi}{3}

\item

\underline{Model explosion}. Today’s LM’s zero-shot performance trails that of fine-tuned models. One can fine-tune the LM to a specific task, but then lose versatility. Contemporary efforts to mitigate the problem focus on training a huge LM jointly on many sets of curated NLP tasks in a massive multi-task setting (several leading studies reaching $100+$ tasks) \cite{EXT5, wei_instructions, metaICL, T0}. These formidable efforts are effective; the resulting models exhibit versatility and high performance when encountering inputs resembling those of the curated tasks. But the performance of these models on tasks that are not close enough to those included in the curated tasks can significantly deteriorate (for example, perplexity degrades significantly). It is not practical to fine-tune and serve multiple large models. Nor can one further tune a multi-task-trained LM \cite{T0, EXT5, metaICL, wei_instructions} on a new task that hadn't been covered in its training; due to catastrophic forgetting, adding the new task necessitates retraining on the entire task set. Given the cost of training such models \cite{training_cost, sustainableAI, energy}, this is clearly infeasible to do repeatedly.

\end{enumerate}

Despite all these shortcomings, large language models are an essential backbone of any future AI system. So the question is how to have our cake and eat it too, enjoying the benefits of self-supervised deep language models without suffering these drawbacks. The solution we offer takes the form of a flexible architecture dubbed the Modular Reasoning, Knowledge and Language (MRKL, pronounced ``miracle'') system, whose high-level design is depicted below. 
\newline
\begin{center}
    \includegraphics[height=8cm]{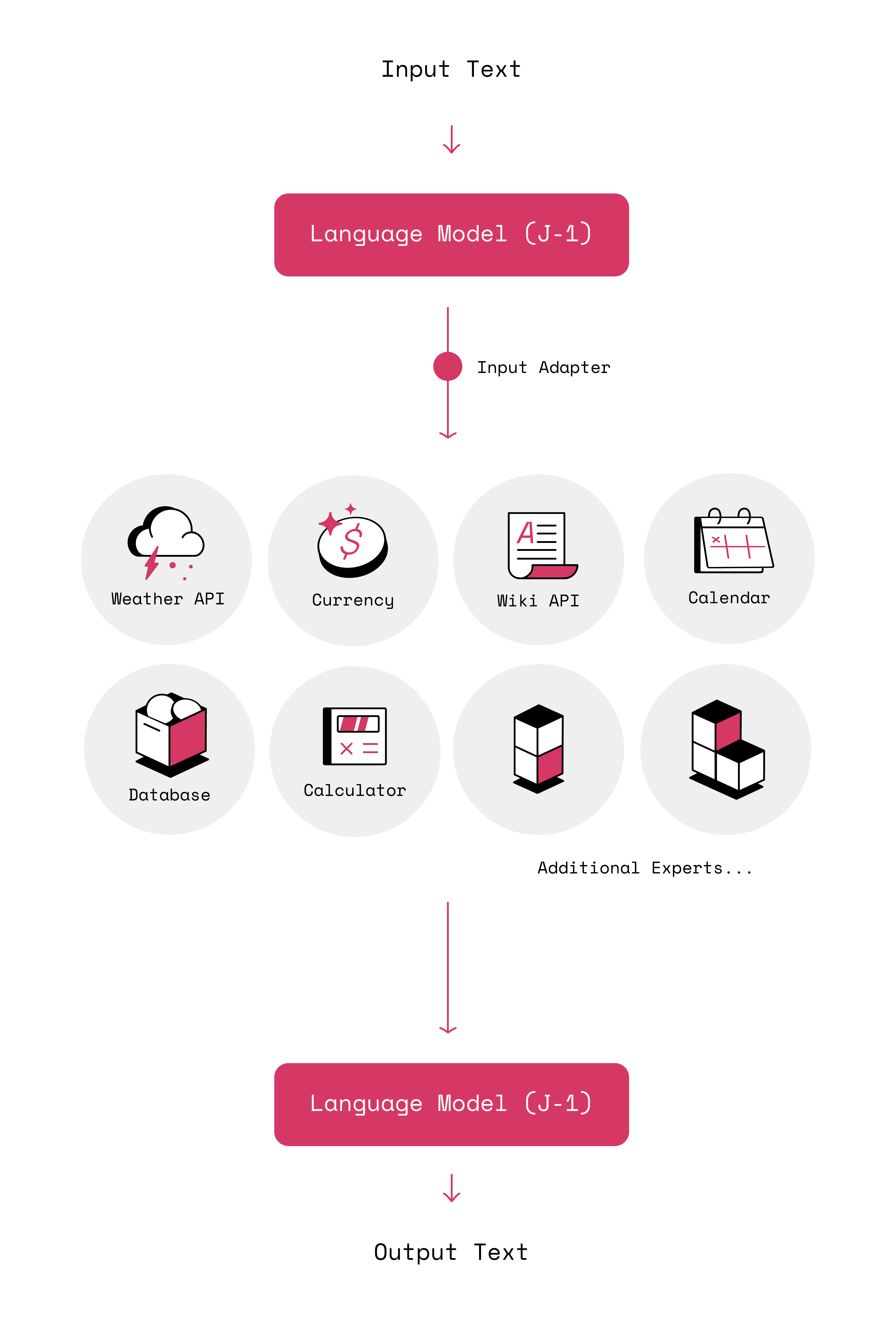}
\end{center}

Thus a MRKL system consists of an extendable set of modules, which we term 'experts', and a router that routes every incoming natural language input to a module that can best respond to the input (the output of that module can be the output of the MRKL system, or be routed to another module). These modules can be:

\begin{itemize}

\item Neural, including the general-purpose huge language model as well as other smaller, specialized LMs.

\item Symbolic, for example a math calculator, a currency converter or an API call to a database.

\end{itemize}

MRKL systems enjoy important benefits when compared to fine-tuned multi-task models:

\begin{enumerate}

    \item Safe fallback: In case the input doesn’t match any existing expert module, the router sends the input directly to the general-purpose huge LM.

\item Robust extensibility: Since each expert is trained independently we are able to cheaply add new capabilities while guaranteeing that they do not compromise the performance of existing ones. The only component that requires retraining is the router which is a relatively lightweight task. 

\item Interpretability: When the router invokes a specific module, that often has the side benefit of providing a rationale for the MRKL system's output (``$1+1=2$ because the calculator said so''); such explanations are crucially lacking in existing language models.

\item Up-to-date information: The integration of external APIs allows the MRKL system to hook into dynamic knowledge bases, and correctly answer inputs that static models cannot.

\item Proprietary knowledge: Access to proprietary databases and other information sources.

\item Compositionality: By routing compounded multi-hop inputs to different experts we are able to naturally integrate their responses and correctly address complex inputs \cite{break}.

\end{enumerate}

\section{Jurassic-X: AI21 Labs’ MRKL system}

We have implemented a MRKL system called Jurassic-X, which is being piloted by a few partners. \\
\\
Jurassic-X will soon be available to developers; you can apply for early access \href{https://dghsiym3kjj.typeform.com/to/rnIzTBQ0}{here}.
\\
\\
Meanwhile, you can experience Jurassic-X via a demo \href{https://studio.ai21.com/jurassic-x}{(link)}.

\section{Crossing the neuro-symbolic chasm: A calculator test case}

There are of course many details involved in implementing a MRKL system. In connection with avoiding model explosion, see our detailed discussion \href{https://storage.cloud.google.com/ai21-public-data/publications/Standing_on_the_shoulders_of_giant_LMs.pdf}{here}.
There is also an interesting challenge of how to intelligently route input among modules, which we leave for a separate discussion. Here we discuss one particularly interesting challenge, that of extracting from the text the formal arguments to symbolic reasoners. 

\subsection{Arithmetic as a test case for chasm crossing}

The relation between neural and symbolic approaches to AI is guaranteed to raise heated discussions about their relative merits with neuro-skeptics on one extreme, neuro-diehards on the other, and the rest trying to have a rational conversation. Before discussing our technical approach we'd like to clarify our guiding ideology. We believe that neural LMs are essential; we didn't build Jurassic-1 \cite{J1} for no reason, and it serves as a backbone to our applications as well as those of developers using AI21 Studio. But as we've made clear, we also believe they have inherent shortcomings. We don't take a firm position on which types of computation are best relegated to the neural machinery, and which should be carved out to symbolic methods. That will surely be an evolving process. What we do firmly believe is that some tasks should be handled by discrete methods, and we should have a clear methodology for how to hand off the computation from the neural net to the symbolic procedure. 

Once the router has made the decision of which module to call upon, it needs to pass the right information to it. The router is a specialized neural net and therefore invoking a neural module is easy since the neural-to-neural interface is natural. However, when a neural network needs to access a database, make an API call, or invoke another symbolic computation, it must extract from the text discrete parameters required by the module. The main message here is that there is no free lunch, but that lunch is in many cases affordable. The cost is in training the router to extract the arguments reliably, which must be done rigorously. 
We make the point by discussing how we trained Jurassic-X to extract basic arithmetic operations.

There are many ways to describe in language a situation that calls for performing arithmetic. The most straightforward way is to use mathematical notation, say $``3-1=?''$. Google search handles it well:
\newline
\begin{center}
    \includegraphics[width=7cm]{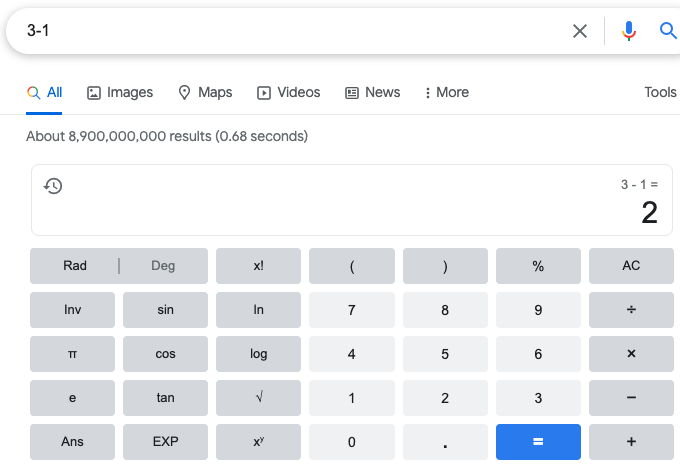}
\end{center}
Not much harder is to express it in words, as in ``How much is three minus 1''. Google search does this well too:
\newline
\begin{center}
    \includegraphics[width=7cm]{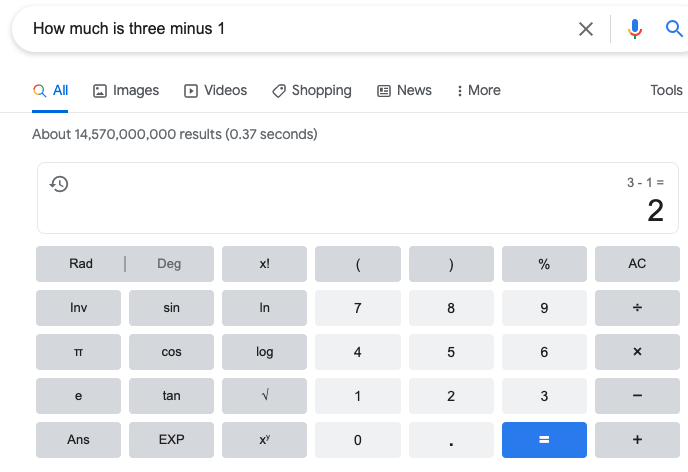}
\end{center}

But then it gets tricky. At the lexical level, one may use different synonyms that carry the same meaning: (``twelve'', ``12'' and ``a dozen''). But beyond simple lexical issues, there are phrases that require world knowledge to understand that there is  an arithmetic exercise encoded in the text (``I lost one ball’’, ``I dropped one ball’’, ``One ball was taken from me’’) and what the specific exercise is. Here Google search starts to stumble:
\newline
\begin{center}
    \includegraphics[width=8cm]{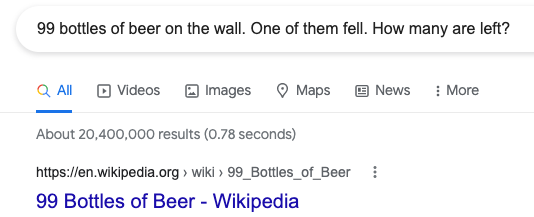}
\end{center}

This is where language models come in. Note that we limit their role to extracting the right arguments for the symbolic calculator, not performing the calculation itself. But we should approach even this task with sober expectations. As anyone knows from elementary school, it's  harder to teach children to solve verbal math problems than to solve explicitly stated math problems. It’s harder for adults too, and it's no different for the computer. At some point the machine will stumble (indeed, at some point humans will disagree about the underlying math problem). But when we decide to invoke a symbolic method we should expect reasonable robustness. 

The solution is to train the router to extract the right input independently for each module, with rigorous evaluation of the performance. By way of example, here is how it was achieved simple, one- and two-operation arithmetics.

\subsection{Training Jurassic-X to extract the arguments for basic arithmetic}

Our goal here is not to show how we can extract the most complex mathematical expressions from text, but how we can extract simple expressions with high reliability, the sort of reliability one would need in a production-grade system. We found that while extensive pretraining does allow Jurassic-1 to extract the arithmetic operation in many cases in a few-shot setting, the performance is far from perfect. See Appendix \ref{sec:few-shot} for a detailed comparison. Using a data augmentation methodology, based on generating examples from a structured example space, we show that the static pretrained LM can be made to achieve near perfect performance. 

We empirically investigate two related research questions:

\begin{enumerate}
    \item 
        What is required in terms of data augmentation to achieve near-perfect performance in the case of basic mathematical operations? 
    \item 
        How well can Jurassic-X generalize between math problems that vary in type or complexity? 
\end{enumerate}

To answer these questions we evaluate the model's performance on cases where the test data is drawn from the same distribution as the training data, as well as on out-of-distribution test data (allowing us, for example, to train a model on addition problems and test it on multiplication problems).

\subsubsection{Data Augmentation}

We use a small set of templates to generate a set of training and test examples. Each example is characterized by the following dimensions:

\begin{enumerate}
    \item The numbers used in the arithmetic expression, henceforth the operands. The operands may be digits (e.g., $48$) or numbers in word form (e.g., forty eight).
    \item The number of digits in the operands. We explore numbers of $1$--$9$ digits. 
     \item The type of operation. We explore addition, subtraction, multiplication and division.
    \item The number of operations in the arithmetic expression. We explore the cases of one and two operations. 
    \item The placement or absence of brackets (in expressions with two operations). In some expressions the placement of brackets may change the result. We explore all logical possibilities for the case of two operations. 
\end{enumerate}

 Table \ref{tab:formats} in the appendix details the templates we use in this set of experiments for the case of a single operation.
In the case of two operations, there are $29$ distinct combinations that have a natural formulation in natural language (see Table \ref{tab:formats-two-ops}). The data can be found \href{https://github.com/AI21Labs/MRKL_synthetic_data}{here}.

\subsubsection{Experimental setup}

We  conducted our experiments with the $7B$ parameters J1-large model \cite{J1} using prompt-tuning \cite{prompt-tune} with $10$ prompt tokens. Across all experiments, we  set our learning rate to be $lr=0.3$ and used linear decay. The batch size was set to $32$ and we  defined a short warm-up depending on the number of steps in each experiment.

Experiments 1 and 2 below were trained for 3000 steps, and the reported results here were the test accuracy evaluated on the final model.
For the remaining experiments we used linear weight decay ($0.001$), which we found to be crucial for the model's performance, and selected the best checkpoint using a validation set. Each experiment was run $3$-$5$ times, and results show mean $\pm$ standard deviation across these runs.

In all experiments we verified that there was no overlap between the problems included in the training set and those included in the test set. This also includes avoiding cases with the same underlying arithmetic expression, but using different wordings for training and testing. For a detailed description of the sizes of the data splits see Section \ref{setup_details}.

\subsubsection{Results}

In the following results we report  the accuracy we achieve in different experiments, by which we mean the percent of problems in the relevant test set on which the system gave the correct answer. (We note again that, since the actual calculation is done by the calculator module, all errors are due to passing the wrong operations or operands to the calculator.)

\paragraph{Experiment 1: Generalization across  different  number of digits in the operands.}

Inspired by the experiments reported by \cite{GPT3}, we test Jurassic-X's ability to generalize to numbers it has not seen in the training data, including numbers with a different (and much larger) number of digits. 
At training time we expose the model only to single-digit numbers, while at test time we evaluate on numbers with $1$ to $9$ digits.
In this section we experiment with simple problems, involving only numbers written as digits (and not numbers as words), one format (format 0 from Table~\ref{tab:formats}), and a single operation. We explore two settings: one where that operation is addition, and one where it is multiplication. 

Table \ref{tab:num_digits} presents our results, sliced by the number of digits, and compared to the results by GPT-3's approach \cite{GPT3} (as representative of all language models that don't have access to external calculator), reported on addition. We note that we trained on single-digit operations for all settings while GPT-3 was conditioned on examples with the same number of digits when answering a certain problem. 
Our results show that despite the fact that training was only done on numbers with a single digit, the model is able to generalize to all numbers of digits explored. This is in stark contrast to the approach of language models which attempt to synthesize arithmetic capabilities from the training data, and as a result  display a dramatic decrease in performance as the number of digits increases. 

\begin{table}[h]
    \centering
    \begin{tabular}{l r r r r r r r r r r}
    \toprule
    \bf Num. digits & \bf 1 & \bf 2 & \bf 3 & \bf 4 & \bf 5 & \bf 6 & \bf 7 & \bf 8 & \bf 9\\ 
    \bf Addition & 1.0 & 1.0 & 1.0 & 1.0 & 1.0 & 1.0 & 1.0 & 1.0 & 1.0\\
    \bf Multiplication & 1.0 & 1.0 & 1.0 & 1.0 & 1.0 & 0.98 & 1.0 & 1.0 & 1.0\\
    \bf GPT-3 & N/A & 1.0 & 0.804 & 0.255 & 0.093 & N/A & N/A & N/A & N/A\\

    \bottomrule 
    \end{tabular}
   	\caption{ {\small \textbf{Robustness to the number of digits in the operands.} Accuracy vs. the number of digits in the test data for a model that was trained only on $1$-digit operands ($5+3$, etc.). Results for GPT-3 were taken from \cite{GPT3}.}}
    \label{tab:num_digits}
\end{table}

\paragraph{Experiment 2: Generalization from digits to numbers as words and vice versa.}

In this set of experiments, we test the model's ability to generalize from arithmetic questions with digits to ones where the numbers are expressed with words, and vice versa. We note that this task is not trivial considering the fact that the tokenization of a number represented with digits is different than that when the same number is represented as words.
For example, we train the model with examples such as ``How much is $58$ plus $12$'' and evaluate it on examples such as ``How much is twenty seven plus thirteen'', and vice versa. 
While we vary the lexical choice for specifying numbers using words or digits, we hold other dimensions of the arithmetic world fixed, including using only format 0 and holding the type of operation fixed at training and test time. (In the following paragraphs we examine these other dimensions.) 

Table \ref{tab:digits_to_words} shows the accuracy for all combinations of training and testing on digits and on words. We notice that in-distribution performance (training and testing on the same type of input) is close to 100 \% accuracy. We further observe that training on words generalizes well to digits. Interestingly, the inverse is not true and the performance is much lower, indicating the underlying difference in the representations of the numbers.

\begin{table}[h]
    \centering
    \begin{tabular}{l r r }
    \toprule
    & \multicolumn{2}{c}{\bf Test} \\ 
    \cmidrule(lr){2-3}
    \bf Train  & \bf Digits & \bf Words \\ 
    \cmidrule(lr){1-1}
    \bf Digits & 1.0 & 0.156 \\
    \bf Words & 0.987 & 0.988 \\
    \bottomrule 
    \end{tabular}
   		\caption{ {\small \textbf{Robustness to wordings.} Accuracy for all combination of training and testing on numbers written as digits and as words. Results are averaged across all the numbers of digits.}}
    \label{tab:digits_to_words}
\end{table}

\paragraph{Experiment 3: Generalization across question formats.}
A main challenge in constructing natural language interfaces to discrete reasoners like a calculator is in handling language variability.
One of the most appealing characteristics of language models is their ability to abstract away from such variability. Here we test the ability of the model to generalize over formats of arithmetic problems.  Experiments are for single-operation problems. 

We train  on a single format (format $0$ in Table \ref{tab:formats}), test on all formats, and break the results by format. We report results with all four operation types. 
Table \ref{tab:exp3} shows that the generalization across formats is perfect in most cases. Format $4$, which is the only one not phrased as a question (see Table \ref{tab:formats}) appears to be the most challenging to generalize to. We note that there are numerous ways to phrase such questions, hence the ability to generalize across formats, even for the case where the model was trained only on one format, is critical for the model's success.

\begin{table}[h]
    \centering
    \begin{tabular}{l r r r r }
    \toprule
    & \bf add & \bf sub & \bf mul & \bf div \\ 
    \bf format 0 & 1.0 $\pm$ 0 & 1.0 $\pm$ 0 & 1.0 $\pm$ 0 & 0.997 $\pm$ 0.006 \\
    \bf format 1 & 1.0 $\pm$ 0 & 1.0 $\pm$ 0 & 1.0 $\pm$ 0 & 1.0 $\pm$ 0 \\
    \bf format 2 & 1.0 $\pm$ 0 & 1.0 $\pm$ 0 & 1.0 $\pm$ 0 & 1.0 $\pm$ 0 \\
    \bf format 3 & 1.0 $\pm$ 0 & 0.993 $\pm$ 0.012 & 0.997 $\pm$ 0.006 & 1.0 $\pm$ 0 \\
    \bf format 4 & 0.993 $\pm$ 0.012 & 0.863 $\pm$ 0.237 & 0.977 $\pm$ 0.04 & 0.727 $\pm$ 0.465 \\
    \bottomrule 
    \end{tabular}
    \caption{\small{\textbf{Robustness to phrasing.} Accuracy when training with one format (format $0$) and evaluating on each of the $5$ formats from Table~\ref{tab:formats}.}}
    \label{tab:exp3}
\end{table}

\paragraph{Experiment 4: Generalization between operations.}

Next we explore whether a model that was trained on one type of arithmetic problems can generalize to other types. We conduct two types of experiments: one where we examine the generalization ability of Jurassic-X on single operation problems, and one with two-operation problems.

For the first set of experiments, we train with one operation at a time, using all formats with numbers as digits. Operands have between $1$ and $9$ digits. We test on all types of single-operation problems. The results are shown in in Table~\ref{tab:exp4a}. Consistent with the previous experiments, training and evaluating on arithmetic problems with the same operations is consistently successful ($accuracy > 0.99$ along the diagonal). Interestingly, we find strong generalization in many cross-operation cases (off the main diagonal in the table). For example, training on addition works almost perfectly when evaluated on subtraction and multiplication. The division operation is an exception, as models trained on it struggle with multiplication and subtraction (but perform reasonably on addition). Conversely, models trained on other operations do not generalize very well to division. 

\begin{table}[h]
    \centering
    \begin{tabular}{l r r r r }
    \toprule 
    & \multicolumn{4}{c}{\bf Test} \\ 
    \cmidrule(lr){2-5}
    \bf Training &  \multicolumn{1}{c}{\bf add} & \multicolumn{1}{c}{\bf sub} & \multicolumn{1}{c}{\bf mul} & \multicolumn{1}{c}{\bf div} \\ 
    \cmidrule(lr){1-1} \cmidrule(lr){2-5} 
    \bf add  & 0.997 $\pm$ 0.006 & 0.96 $\pm$ 0.026 & 0.97 $\pm$ 0.006 & 0.477 $\pm$ 0.127  \\
    \bf sub & 0.987 $\pm$ 0.023 & 0.997 $\pm$ 0.006 & 0.763 $\pm$ 0.182 & 0.183 $\pm$ 0.196 \\
    \bf mul & 0.483 $\pm$ 0.466 & 0.817 $\pm$ 0.186 & 0.993 $\pm$ 0.006 & 0.26 $\pm$ 0.2 \\
    \bf div & 0.787 $\pm$ 0.244 & 0.31 $\pm$ 0.334 & 0.23 $\pm$ 0.214 & 0.993 $\pm$ 0.012 \\
    \bottomrule
    \end{tabular}
    \caption{\small{\textbf{Generalization across operations for single-operation problems.} Accuracy when training with one operation and evaluating on other operations.}}
    \label{tab:exp4a}
\end{table}

For the second set of experiments, we randomly partition the $29$ possible two-operation arithmetic problems (Tables~\ref{tab:two-ops}, \ref{tab:formats-two-ops}) into subsets of $14$ and $15$ to be used for training and testing, respectively. We repeated this process for 10 random splits of the operations, assuring that only one of the training two-operation problems would include brackets. Other settings are as with the first set of experiments.

Table \ref{exp4:2ops} shows the average accuracy for each of the $29$ types of two-operation problems.
The results show that in $22$ of the $29$ combinations the accuracy exceeds $90\%$, indicating reasonable generalization capabilities for such a setting.

\begin{table}[!ht]
    \centering
    \begin{tabular}{|l|l|l|}
    \hline
        \bf Formula & \bf Mean & \bf STD \\ \hline
        f=((A+B)*C) & 0.288 & 0.267 \\ \hline
        f=(A+B*C) & 0.406 & 0.349 \\ \hline
        f=((A-B)*C) & 0.562 & 0.194 \\ \hline
        f=(A/(B/C)) & 0.677 & 0.228 \\ \hline
        f=(A-B*C) & 0.697 & 0.266 \\ \hline
        f=(A*(B-C)) & 0.797 & 0.325 \\ \hline
        f=((A+B)/C) & 0.867 & 0.265 \\ \hline
        f=(A-(B-C)) & 0.930 & 0.057 \\ \hline
        f=((A-B)/C) & 0.940 & 0.120 \\ \hline
        f=(A-B/C) & 0.955 & 0.042 \\ \hline
        f=(A/(B+C)) & 0.960 & 0.042 \\ \hline
        f=(A/(B-C)) & 0.962 & 0.094 \\ \hline
        f=(A+B/C) & 0.964 & 0.049 \\ \hline
        f=(A*(B/C)) & 0.970 & 0.052 \\ \hline
        f=(A*B+C) & 0.973 & 0.025 \\ \hline
        f=(A*(B+C)) & 0.974 & 0.025 \\ \hline
        f=(A/B+C) & 0.981 & 0.024 \\ \hline
        f=(A/B/C) & 0.985 & 0.017 \\ \hline
        f=(A/B-C) & 0.987 & 0.022 \\ \hline
        f=(A/B*C) & 0.99 & 0.015 \\ \hline
        f=(A-(B+C)) & 0.99 & 0.017 \\ \hline
        f=(A*B-C) & 0.992 & 0.015 \\ \hline
        f=(A/(B*C)) & 0.995 & 0.012 \\ \hline
        f=(A-B+C) & 1 & 0 \\ \hline
        f=(A+B+C) & 1 & 0 \\ \hline
        f=(A-B-C) & 1 & 0 \\ \hline
        f=(A*B/C) & 1 & 0 \\ \hline
        f=(A+B-C) & 1 & 0 \\ \hline
        f=(A*B*C) & 1 & 0 \\ \hline
    \end{tabular}
    \caption{ {\small \textbf{Generalization for two-operation problems.} Mean and standard deviation across 10 partitions of the formulae to train and test.}} \label{exp4:2ops}
\end{table}

\paragraph{Experiment 5: Generalization across a different number of operations.}
This final set of experiments explores the ability of the model to generalize from single-operation problems to two-operation problems. 

As above, we use all formats ($0$-$4$), with operands of $1$ to $9$ digits (numbers written with digits), and train on examples from all four operation types (single-operation problems only). We evaluate on examples with two operations, focusing on the situations that do not require bracketing -- $16$ unique combinations in total. The results are shown in Table~\ref{tab:exp5} and are organized by the first operation (rows) and second operation (columns) in the test problems.  

In almost all cases, we observe excellent performance on solving two-operation problems, despite seeing only single-operation problems at training time. This result is also important for the systematic development of the capability since the number of possible formulae grows rapidly with the number of operations, rendering an exhaustive training of all combinations extremely challenging for multi-operation formulae. There are three exceptions of failure to  generalize to two-operation problems: add--mul, sub--mul, and sub--div. In these three cases, the templates are of the form of two prefix phrasings (e.g., ``what is the sum of $2$ and the product of $4$ and $8$''), while other templates are relatively simpler (Table~\ref{tab:formats-two-ops}). However, we obtain pretty good performance on add--div, despite it having a similar template.
 
\begin{table}[h]
    \centering
    \begin{tabular}{l r r r r }
    \toprule 
    & \multicolumn{4}{c}{\bf Second operation} \\ 
    \cmidrule(lr){2-5}
    \bf First operation & \bf add & \bf sub & \bf mul & \bf div \\ 
    \cmidrule(lr){1-1} \cmidrule(lr){2-5} 
    \bf add  & 1.0 $\pm$ 0 & 1.0 $\pm$ 0 & 0.01 $\pm$ 0.017 & 0.784 $\pm$ 0.112  \\
    \bf sub & 0.998 $\pm$ 0.004 & 0.998 $\pm$ 0.004 & 0.202 $\pm$ 0.247 & 0.224 $\pm$ 0.213 \\
    \bf mul & 0.996 $\pm$ 0.005 & 0.974 $\pm$ 0.047 & 0.998 $\pm$ 0.004 & 1.0 $\pm$ 0 \\
    \bf div & 0.93 $\pm$ 0.063 & 0.834 $\pm$ 0.288 & 0.99 $\pm$ 0.012 & 0.946 $\pm$ 0.11 \\
    \bottomrule
    \end{tabular}
    \caption{\small{\textbf{Generalization from one operation to two operations.} Accuracy when training on single-operation problems and testing on two-operation problems, presented for all operation pairs.}}
    \label{tab:exp5}
\end{table}

\section{Discussion}
This paper introduces the concept of Modular Reasoning, Knowledge and Language (MRKL) systems, which embraces large language models (LMs) and augments them with an easily extensible set of external knowledge and reasoning modules. This flexible, neuro-symbolic design retains all the advantages of modern LMs, but avoids their limitations: a lack of current and/or proprietary information, and an inability to reason symbolically when needed. We discussed some of the technical challenges in implementing a MRKL system, with a special focus on how to cross the neuro-symbolic divide, which we did by looking at how Jurassic-X --- AI21 Labs' implementation of a MRKL system --- was trained to handle basic arithmetic reliably.

\appendix
\newpage
\section{Few-shot vs. prompt tuning} \label{sec:few-shot}
Performing these experiments in a  few-shot setting might be a more natural choice that requires less effort in training these models. However, as the analysis below reveals, the performance of this approach is limited, demonstrating the need to follow a systematic approach. 

\paragraph{Set up:} In the few-shot experiments, $10$ examples were drawn from the training sets and used as a prompt for the model. The performance was evaluated on the same test set for both the few-shot and the prompt-tuning approaches.
\paragraph{Changing the number of digits:} Both methods easily generalize to varying the number of digits (see Table \ref{few_shot_num_digits}). 
\begin{table}[h]
    \centering
    \begin{tabular}{l r r r r r r r r r r}
    \toprule
    \bf Num. digits & \bf 1 & \bf 2 & \bf 3 & \bf 4 & \bf 5 & \bf 6 & \bf 7 & \bf 8 & \bf 9\\ 
    \bf Prompt tuning & 1.0 & 1.0 & 1.0 & 1.0 & 1.0 & 1.0 & 1.0 & 1.0 & 1.0\\
    \bf Few-shot & 1.0 & 1.0 & 1.0 & 1.0 & 1.0 & 1.0 & 1.0 & 1.0 & 1.0\\
    \bottomrule 
    \end{tabular}
   	\caption{ {\small \textbf{Number of digits: few-shot vs.\ prompt tuning.} Results shown for addition.}}
    \label{few_shot_num_digits}
\end{table}

\paragraph{Writing the operands in words instead of digits:} A clear difference appears as the few-shot approach is unable to handle this task and the performance greatly decreases with the number of digits (see Table \ref{few_shot_words}).
\begin{table}[h]
    \centering
    \begin{tabular}{l r r r r r r r r r r}
    \toprule
    \bf Num. digits & \bf 1 & \bf 2 & \bf 3 & \bf 4 & \bf 5 & \bf 6 & \bf 7 & \bf 8 & \bf 9\\ 
    \bf Prompt tuning & 1.0 & 1.0 & 1.0 & 0.95 & 0.99 & 1.0 & 0.98 & 0.99 & 0.98\\
    \bf Few-shot & 1.0 & 1.0 & 0.93 & 0.62 & 0.6 & 0.38 & 0.17 & 0.03 & 0.0\\
    \bottomrule 
    \end{tabular}
   	\caption{ {\small \textbf{Number of digits: few-shot vs.\ prompt tuning, when numbers are written as words.} Results are shown for addition.}}
    \label{few_shot_words}
\end{table}

\paragraph{Different question formats:} We explored the ability to generalize to different question formats when training on one format (format $0$). We note the generalization is much lower compared to following our prompt-tuning method (Table \ref{few_shot_formats}).
\begin{table}[h]
    \centering
    \begin{tabular}{l r r r r r r}
    \toprule
    \bf Num. digits & \bf Format 0 & \bf Format 1 & \bf Format 2 & \bf Format 3 & \bf Format 4\\ 
    \bf Prompt tuning & 1.0 & 1.0 & 1.0 & 1.0 & 0.993 \\
    \bf Few-shot & 1.0 & 1.0 & 0.98 & 0.78 & 0.22 \\
    \bottomrule 
    \end{tabular}
   	\caption{ {\small \textbf{Generalization across formats: few-shot vs. prompt tuning.} Results are shown for addition.}}
    \label{few_shot_formats}
\end{table}

\section {Question formats} \label{two-op formats}
Table \ref{tab:formats} shows the five formats used for single-operation problems for all operations. 
 \begin{table}[h]
\small 
    \centering
    \begin{tabular}{c ll}
    \toprule
    \textbf{Format} & \multicolumn{1}{c}{\bf addition} & \multicolumn{1}{c}{\bf subtraction}   \\ 
    0 & How much is \{x\} plus \{y\}? & How much is \{x\} minus \{y\}?  \\
    1 & What is \{x\} plus \{y\}? & What is \{x\} minus \{y\}?  \\
    2 & What is the result of \{x\} plus \{y\}? & What is the result of \{x\} minus \{y\}? \\ 
    3 & What is the sum of \{x\} and \{y\}? & What is the difference between \{x\} and \{y\}? \\ 
    4 & The sum of \{x\} and \{y\} is & The difference between \{x\} and \{y\} is \\
    \midrule 
    & \multicolumn{1}{c}{\bf multiplication}  & \multicolumn{1}{c}{\bf division} \\ 
    0 & How much is \{x\} times \{y\}? & How much is \{x\} over \{y\}? \\ 
    1 & What is \{x\} times \{y\}? & What is \{x\} over \{y\}? \\  
    2 & What is the result of \{x\} times \{y\}? & What is the result of \{x\} over \{y\}?  \\  
    3 & What is the product of \{x\} and \{y\}? & What is the ratio between \{x\} and \{y\}? \\  
    4 & The product of \{x\} and \{y\} is & The ratio of \{x\} and \{y\} is \\
    \bottomrule
    \end{tabular}
    \caption{\small{\textbf{Question formats for single-operation problems.} Five different formats for each type of operation in the case of single-operation expressions.}}
    \label{tab:formats}
\end{table}

The equations for two-operation problems, for all combinations of placing the brackets, are shown in Table \ref{tab:two-ops}. Table \ref{tab:formats-two-ops} shows the phrasings for these $29$ formulae.

\begin{table}[t]
    \centering
    \begin{tabular}{l | r|r|r|r}
    \toprule 
    & \multicolumn{1}{c}{$+$} & \multicolumn{1}{c}{$-$} & \multicolumn{1}{c}{$*$} & \multicolumn{1}{c}{$/$} \\ 
    \midrule 
    $+$ &  \makecell{$A+B+C$} & \makecell{$A+B-C$} & \makecell{$(A+B)*C$ \\ $A+(B*C)$  } & \makecell{$(A+B)/C$ \\ $A+(B/C)$} \\ 
    \midrule 
    $-$ & \makecell{ $(A-B)+C$ \\ $A-(B+C)$} & \makecell{ $(A-B)-C$ \\ $A-(B-C)$} & \makecell{$(A-B)*C$ \\ $A-(B*C)$} & \makecell{ $(A-B)/C$ \\ $A-(B/C)$} \\  
    \midrule
    $*$ & \makecell{ $A*(B+C)$ \\ $(A*B)+C$} & \makecell{ $A*(B-C)$ \\ $(A*B)-C$} & \makecell{$A*B*C$} & \makecell{$(A*B)/C$ \\ $A*(B/C)$} \\ 
    \midrule
    $/$ & \makecell{$ A/(B+C)$ \\ $(A/B)+C$} & \makecell{$A/(B-C)$ \\ $(A/B)-C$} & \makecell{ $(A/B)*C$ \\ $A/(B*C)$} & \makecell{ $(A/B)/C$ \\ $A/(B/C)$} \\ 
    \bottomrule 
    \end{tabular}
    \caption{\small{\textbf{Two-operation formulae.} All combinations of two-operation formulae, including operator precedence where relevant. See Table \ref{tab:formats-two-ops} for the phrasings corresponding to these formulae.}}
    \label{tab:two-ops}
\end{table}

\begin{table}[!ht]
    \centering
    \begin{tabular}{|l|l|}
    \hline
        \bf Formula & \bf Format \\ \hline
        f=((A+B)*C) & Sum A and B and multiply by C \\ \hline
        f=(A+B*C) & What is the sum of A and the product of B and C? \\ \hline
        f=((A-B)*C) & What is the product of A minus B and C? \\ \hline
        f=(A/(B/C)) & How much is A divided by the ratio between B and C? \\ \hline
        f=(A-B*C) & What is the difference between A and the product of B and C? \\ \hline
        f=(A*(B-C)) & How much is A times the difference between B and C? \\ \hline
        f=((A+B)/C) & What is the ratio between A plus B and C? \\ \hline
        f=(A-(B-C)) & How much is A minus the diffrence between B and C? \\ \hline
        f=((A-B)/C) & What is the ratio between A minus B and C? \\ \hline
        f=(A-B/C) & What is the difference between A and the ratio between B and C? \\ \hline
        f=(A/(B+C)) & How much is A divided bu the sum of B and C? \\ \hline
        f=(A/(B-C)) & How much is A divided by the difference between B and C? \\ \hline
        f=(A+B/C) & what is the sum of A and the ratio between B and C? \\ \hline
        f=(A*(B/C)) & How much is A times the ratio between B and C? \\ \hline
        f=(A*B+C) & How much is the sum of A times B and C? \\ \hline
        f=(A*(B+C)) & How much is A times the sum of B and C? \\ \hline
        f=(A/B+C) & How much is the sum of A divided by B and C? \\ \hline
        f=(A/B/C) & How much is A divided by B divided by C? \\ \hline
        f=(A/B-C) & How much is the difference between A divided by B and C? \\ \hline
        f=(A/B*C) & How much is A divided by B times C? \\ \hline
        f=(A-(B+C)) & How much is A minus the sum of B and C? \\ \hline
        f=(A*B-C) & How much is the difference between A times B and C? \\ \hline
        f=(A/(B*C)) & How much is A divided by the product of B and C? \\ \hline
        f=(A-B+C) & How much is A minus B plus C? \\ \hline
        f=(A+B+C) & How much is A plus B plus C? \\ \hline
        f=(A-B-C) & How much is A minus B minus C? \\ \hline
        f=(A*B/C) & How much is A times B divided by C? \\ \hline
        f=(A+B-C) & How much is A plus B minus C? \\ \hline
        f=(A*B*C) & How much is A times B times C? \\ \hline
    \end{tabular}
    	\caption{ {\small \textbf{Formats of two-operation questions.}}} \label{tab:formats-two-ops}
\end{table}

\section{Amount of data in each experiment} \label{setup_details}
\textbf{Experiment $1$ - number of digits:}
\newline
The training set included $40$ single digits operand combinations. The test set for single digits included the remaining $41$ combinations (discarding $0$) and $50$ randomly chosen combinations for all the cases with more then $1$ digit.
\newline
\textbf{Experiment $2$ - digits and words:}
\newline
$200$ samples were randomly drawn for each number of digits (except for $1$ and $2$ digits) and data was split equally between train and test data. This was repeated for both digits and words. 
\newline
\textbf{Experiment $3$ - question formats:}
\newline
The train set included $400$ samples with format $0$ and the dev and test set each included $200$ samples from each format, randomly drawn with number of digits ranging between $1$ and $9$.
\newline
\textbf{Experiment 4 - generalization across operations:}
\newline
The train set included $635$ samples across all formats and digits for each operation, while the dev and test set each included $315$ samples for each operation.
\newline
For the two-operation experiment we drew $120$ samples for each of the $29$ formats, which were divided equally between the train, dev and test sets.
\newline
\textbf{Experiment 5 - generalization across the number of operations:}
\newline
$700$ samples with a single operation were used for training the model, drawn randomly from all operations, formats, and number of digits.
The dev and test set each included 210 samples drawn randomly for each of the $16$ combinations of two operation formulas, analyzed up to 7 digits.

\clearpage

\printbibliography

\end{document}